\documentclass[letterpaper, 10 pt, conference]{ieeeconf}  

\usepackage{cite}
\usepackage{graphicx, amsmath, amssymb} 
\usepackage{array}
\usepackage{algorithm}
\usepackage{algorithmic}
\usepackage[utf8]{inputenc}
\usepackage{tikz}
\usetikzlibrary{shapes.geometric, arrows, positioning}
\usepackage{balance}
\makeatletter
\let\NAT@parse\undefined
\makeatother
\usepackage[hidelinks]{hyperref}

\IEEEoverridecommandlockouts
\usepackage{fancyhdr}
\fancypagestyle{firstpage}{
  \fancyhf{}
  
  \fancyfoot[C]{\footnotesize Accepted at the 2024 10th International Conference on Control, Decision and Information Technologies (CoDIT)}
}

\title{\LARGE \bf
Primitive Agentic First-Order Optimization
}
\author{R. Sala$^{1}$ 
\thanks{*This work was supported by the Deutsche Forschungsgemeinschaft (DFG) Project ID 465355776.}
\thanks{$^{1}$Department of Mechanical and Process Engineering, 
        RPTU Kaiserslautern, Germany 
        {\tt\small r.sala@rptu.de}}%
}

\begin{document}

\maketitle
\thispagestyle{firstpage}

\begin{abstract}
Efficient numerical optimization methods can improve performance and reduce the environmental impact of computing in many applications.
This work presents a proof-of-concept study combining primitive state representations and agent-environment interactions as first-order optimizers in the setting of budget-limited optimization. 
Through reinforcement learning (RL) over a set of training instances of an optimization problem class, optimal 
policies for sequential update selection of algorithmic iteration steps are approximated in generally formulated low-dimensional partial state representations that consider aspects of progress and resource use.
For the investigated case studies, deployment of the trained agents to unseen instances of the quadratic 
optimization problem classes outperformed conventional optimal algorithms with optimized hyperparameters. The results show that elementary RL methods combined with succinct partial state representations can be used as heuristics to manage complexity in RL-based optimization, paving the way for agentic optimization approaches. 
\end{abstract}


\section{INTRODUCTION AND MOTIVATION}
Numerical optimization methods are essential in the realization of many modern artificial intelligence and engineering systems. 
Progress in optimization efficiency is relevant not only for large-scale applications, such as scientific computing on supercomputer clusters, but can also lead to benefits on smaller scales, such as increased performance or extended battery life on smartwatches, phones, and small autonomous robotic systems.  

Although there is a great variety of sophisticated optimization algorithms and variants,
relatively straightforward gradient-based algorithms such as Stochastic Gradient Descend (SGD) and Nesterov Accelerated Gradient descend (NAG) \cite{Nesterov1983} are still widely used in first-order settings where only function evaluation values and gradients are available. NAG is theoretically well-suited for optimization problems that have smooth and strongly convex objective functions. However, in practice, it is also often used beyond these restrictive requirements. Popular applications range from non-convex neural network training \cite{sutskever2013importance}, control, robotics \cite{wensing2023optimization}, to topology optimization \cite{oka23nesterov}, and engineering design \cite{liu2017survey}. Due to the widespread and intensive use of first-order optimization methods, even small improvements can lead to substantial overall reductions in resource use. 

The importance of selecting algorithms and hyper-parameters has been emphasized in the context of convex optimization \cite{agrawal2020learning},  neural network training \cite{domhan2015speeding,sutskever2013importance,yang2020hyperparameter}, and non-convex optimization in general \cite{dauphin2015equilibrated,
guo2019new,
sala20benchmarking}. Where recent advancements focused on automation of \textit{a priori} algorithm and parameter selection on a high-level (per run)
\cite{
meunier21black,wu2023selecting}, this work targets these aspects on the level of the algorithmic updates in each iteration. By breaking down the task into multiple steps with various possible actions and outcomes, computational rationality 
is introduced at a finer-grained level within the optimization process. 
The aim is to improve optimization efficiency by combining agentic aspects with succinct partial state representations that consider features of progress toward the objective as well as resource use.

\section{Related Work}
This literature overview focuses on developments related to accelerated gradient first-order optimization methods, Agent-based, and RL-based optimization. The aim is to connect these research areas to the topic of Primitive Agentic RL-based first-order optimization. 

\subsection{Accelerated Gradient Methods}
In \cite{Nesterov1983} NAG was presented with the proof that for first-order optimization of smooth and strongly convex functions NAG achieves the best possible (optimal) accuracy convergence rate of $O(1/K^2)$, where $K$ is the number of iterations.
The relation between NAG and earlier classical momentum (CM)  methods \cite{polyak1964some} was analyzed in \cite{sutskever2013importance}.
Further insights into the convergence behavior of NAG and other gradient-based optimization methods were obtained by relating them to continuous-time ordinary differential equations \cite{su2016differential, sala2022euler}. This inspired the formulation of a family of generalized Nesterov schemes with similar optimal quadratic local convergence rates~\cite{su2016differential}.
Although the advantages of the optimal local convergence rate of NAG for full batch gradients is lost in the stochastic gradient estimation setting, NAG and similar algorithms can still perform very well in the 'primary' or 'transient phase' before local convergence sets in \cite{sutskever2013importance}. This transient phase is important in training deep neural networks and in the more general setting of first-order optimization under computational resource budget limitations.

The trade-off between progress in the transient phase and the local convergence 'endgame' co-motivated the development of stepsize or learning rate decay and other modifications. Mathematical interpretations of the benefits of learning rate decay for stochastic gradient algorithms are given in \cite{shi2023learning}. Some challenges and advantages of geometric decay over polynomial decay for fixed time-horizon stochastic gradient procedures were analyzed in \cite{ge2019step}. The advantages of cyclic learning rates over fixed and exponential decay for some applications were presented in \cite{smith2017cyclical}.
Results of \cite{smith2017don} indicated that in some settings increasing the batch size instead of using learning rate decay can be beneficial. 


\subsection{Reinforcement Learning based Optimization}
In the scope of this work, optimizing first-order optimization problems is considered from an RL perspective. Although there is much literature on optimization for RL, there are far fewer publications that explicitly address RL-related approaches for the optimization setting. 

Fundamental ideas and theories on learning and adaptation were described in \cite{holland1975adaptation,vapnik2006estimation}, while the foundations of RL were described in \cite{bellman57dynamic, Sutton98}. The idea of RL-based training of agents for function optimization is also related to the general concept of meta-learning  \cite{vilalta2002perspective,vanschoren2019meta}. An early application of the meta-idea to apply learning to learning rules was presented in \cite{bengio1991learning}. 
Another early approach pointed implicitly in the direction of applying learning methods to optimization, in the context of learning fast approximations for sparse coding \cite{gregor2010learning}. 
In \cite{zaremba2016learning} the ability of neural networks to learn simple algorithms in a supervised setting as well as in a Q-learning setting was investigated. Later work that more explicitly applied learning approaches to more general optimization problems was presented in  \cite{andrychowicz2016learning}. More specifically, the work in \cite{andrychowicz2016learning} demonstrated the use of Long Short-Term Memory (LSTM) recurrent neural networks that were trained as optimizers for quadratic functions and loss functions for neural network training. 

A more RL-oriented perspective towards function optimization was initiated as 'learning to optimize' (L2O) presented in \cite{li2016learning}. 
There, a neural network was used to represent the agents' policy about the direction and stepsize for each optimization iteration. The agent's state representation was based on a moving window with information from the previous iterations, such as location in the domain, gradients, and objective value changes.
Only relatively recently, research in this explicit RL-based optimization direction has started to expand. 
In \cite{solozabal2019virtual}, a neural network-based RL model was developed and used as a heuristic for NP-hard combinatorial network function placement optimization problems. 
A multi-agent-based deep RL approach for energy grid control optimization was described in \cite{zhang2023multi}. 
An agent-based collaborative random search method was presented and compared with two common randomized tuning strategies for hyper-parameter tuning in the setting of regression, classification, and approximate global function optimization in \cite{esmaeili2023agent}. 
An evolutionary RL-based approach for hyperparameter optimization in classification tasks was introduced in \cite{shin2022evolutionary}.


One closely related work is \cite{wu2023selecting} in which a recommendation system for optimized learning rates for common datasets was presented. Although the term "policies" is used in \cite{wu2023selecting}, it refers there to configurations of parameters or dynamic functions for learning rates, rather than the agent policies that encompass a spectrum of actions over a state space, as in the RL framework. In contrast, the architecture of their approach integrates a database along with optimization algorithms and a recommendation system to 
select and compose effective learning rate strategies, with considerable efficiency gains compared to common conventional settings. 

Due to the high scientific and industrial interest in efficient optimization, improvements by 'low hanging fruits' are becoming rarer, and most progress seems to come from advanced and relatively complex approaches (e.g. \cite{wu2023selecting,shen2024efficient, chen2022learning, mallik2024priorband}).
A complementary trend to improve optimization efficiency is the development of specialized approaches targeting particular important applications and problem classes \cite{wang2024dp, wang2021reliability}.
Naively, transforming an optimization problem into an RL problem can just make it more complex, less computationally tractable, and less practically useful.  
RL-problems can be subjected to the 'curse of dimensionality'~\cite{bellman57dynamic,Sutton98} and despite recent advancements \cite{li24settling} finding approximately optimal or useful policies can be computationally expensive as the complexity increases with the dimension of the state space, action space and horizon length. However, combined with proper assumptions, structure, simplifications, or heuristics, several useful RL-based optimization approaches have been developed \cite{solozabal2019virtual,zhang2023multi,esmaeili2023agent}. 
This work aims to find useful and accessible optimization efficiency improvements using basic RL methods combined with primitive or succinct low-dimensional state representations. Relatively simple methods have the benefit that they can be communicated and implemented more easily. Moreover, simple and fast methods could also be deployed on small, low-end hardware in settings with strong computational restrictions or time-critical applications such as control systems.   

\section{Problem setting and Methods}
This paper presents a proof-of-concept investigation into a reinforcement learning-based optimization method for continuous unconstrained (convex) smooth first-order optimization, with the additional restriction of a fixed function evaluation budget $K$. 
In this context, the conventional aim is to minimize objective functions $f: \mathbb{R}^n \rightarrow \mathbb{R}$ and to identify the corresponding optimal solution $x^*$. However, in the context of finite resources, here modeled via a finite iteration budget $K$, it may be infeasible to find solutions that are arbitrarily close to the exact (global) minimum. 

\subsection{Conventional Accelerated Gradient Methods}
The original description of the NAG algorithm \cite{Nesterov1983}, contains an implicit step-size condition to adapt the learning rate, and except for two arbitrary initialized points in the domain that effectively determine the initial learning rate, there are no free algorithm parameters. 
As noted in the appendix of  \cite{sutskever2013importance} for a known value of the 
Lipschitz constant $L$ of the gradient $\nabla f$ of the objective function $f$, a fixed learning rate $\eta_k=1/L$ would be sufficient to obtain optimal convergence. 
In \cite{sutskever2013importance}
it is also shown that the iterative scheme of 
NAG, with a fixed learning rate and for large $k$ can be approximated as
\begin{equation}
\begin{split}
\upsilon_{k+1} &= \mu \upsilon_{k} - \eta \nabla f(x_{k} + \mu \upsilon_{k}), \\
x_{k+1} &= x_{k} + \upsilon_{k+1},
\label{eq:NAG}
\end{split}
\end{equation}
starting from a given trial vector $x_1$, 'velocity' vector $v_1 = \mathbf{0}$ for a given 
'momentum' parameter $\mu$, and learning rate $\eta$.
Although not completely equivalent to the original algorithm, this approach has become a widely used way of explaining NAG in the literature and implemented in many software packages. 
For convenience, an additional variable or 'ad interim' point $\Tilde{x_k} = x_k+mv_k $ at which the gradients are evaluated is often introduced. 
Instead of a constant step size $\eta$, various types of learning rate schedules, such as time-based, polynomial, and exponential decay, can be used. In this study exponential decay is considered as an example
\begin{align}
\eta_k &= \eta_1 \cdot e^{-\delta k}. \label{eq:exp_decay}
\end{align}

\subsection{Agent-Environment based Sequential Update Selection}
This subsection describes a novel approach that casts the iterative optimization process into the perspective of a sequence of agent-environment interactions with primitive low-dimensional partial state representations. 
Similarly, as with systems design and other agent-based systems, there could be multiple ways to partition the aspects of a process. The remainder of this section describes one possible paradigm that is investigated in this work. 

In order to transition from the conventional optimization perspective to a formulation that can fit Partially Observable Markov Decision Processes (POMDP), and the Agent-Environment interaction theme in the framework of RL, it can be useful to disentangle and segregate the updates in equation (\ref{eq:NAG}) in steps that can be separated in two stages preceded by an initialization. 
Many first-order optimization algorithms can be more generally described by the following procedure:
Given parameters $\theta$, initialize $x_1$ arbitrarily, then for $k=1, \ldots, K$ repeat the updates: 
\begin{align}
y_k &= f(x_k), \label{eq:yk} \\
g_k &= \nabla f(x_k), \label{eq:gk} \\
x_{k+1} &= H(x_k, y_k, g_k, k,\theta), \label{eq:xkp1} \\
x_k &= x_{k+1}. \label{eq:xk}
\end{align}
Where $H()$ represents a first-order optimization update that determines the next design variable evaluation point $x_{k+1}$, based on the previous evaluation point $x_k$, the corresponding function and gradient evaluation values $f_k,~g_k$ and a tuple of parameters $\theta$ (e.g., $\theta= (\eta, \mu, \delta)$). Examples of such updates can be based on existing algorithms and update rules, such as those of Gradient Descend (GD), NAG, or even a new arbitrary random point. As an example, Algorithm \ref{al:NAGU} provides the pseudo-code of a corresponding NAG-update with exponential decay. 

\begin{algorithm}
\caption{NAG-Update function $H(\Tilde{x_k},g_k,k,\theta$)} \label{al:NAGU}
\begin{algorithmic}
\STATE \textbf{Input:} $\Tilde{x}_{k}, g_k, k, \theta = (\eta, \mu, \delta)$
\STATE \textbf{Output:} $\Tilde{x}_{k+1}$
\STATE \textbf{Persistent:} $x_k,v_k$ (Retain across calls)

If $k == 1$ Then $v_1 \gets 0, x_1 \gets, \Tilde{x_1}$ EndIf
\STATE Determine learning rate decay, $\eta_{k} = \eta \cdot \exp(-\delta \cdot k)$
\STATE Update velocity $v_{k+1} = \mu \cdot v_k - \eta_k \cdot g_k$
\STATE Update position $x_{k+1} = x_k + v_{k+1}$
\STATE Compute interim update $\Tilde{x}_{k+1} = x_k + \mu \cdot v_{k+1}$
\STATE Update persistent $x_k \gets x_{k+1},v_k \gets v_{k+1}$ 
\STATE Return $\Tilde{x}_{k+1}$
\end{algorithmic}
\end{algorithm}

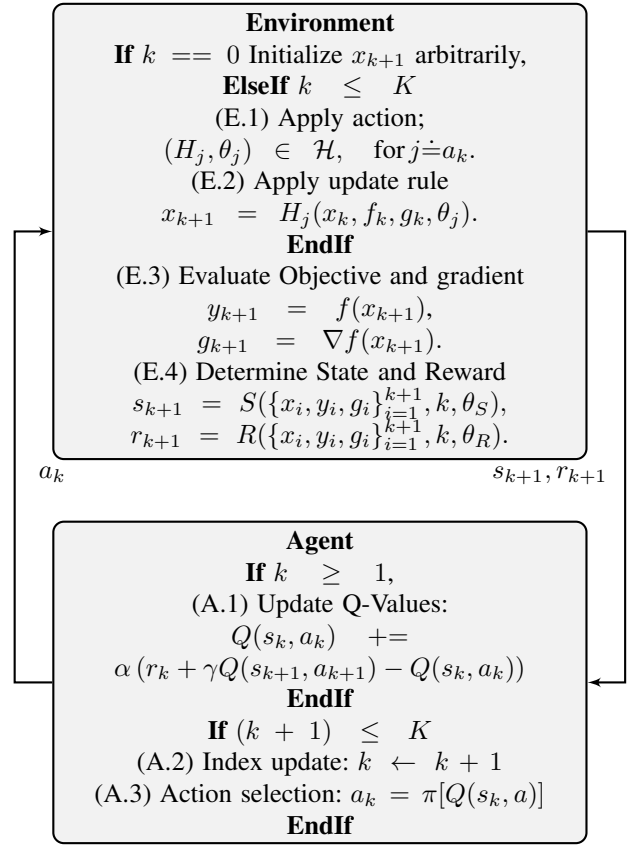
\begin{figure}  
    \centering
    \begin{tikzpicture}[>=stealth', auto, node distance=6.cm, thick]
        \tikzstyle{block} = [rectangle, draw, fill=gray!10, 
            text width=19.5em, text centered, rounded corners, minimum height=3.9em]
        \tikzstyle{line} = [draw, -latex']
        
        \newcommand{\horizlineextend}{1cm} 

        \node [block] (environment) {\textbf{Environment} \\ 
        \textbf{If} $k==0$ Initialize $x_{k+1}$ arbitrarily, \\
        \textbf{ElseIf} $k \leq K$\\ 
        (E.1) Apply action;  \\
        $ (H_j,\theta_j) \in \mathcal{H}, \quad \text{for} \, j \dot{=} a_k. 
        $\\
        (E.2) Apply update rule \\
        $x_{k+1} = H_{j}(x_{k}, f_k, g_k,\theta_{j}).$\\
        \textbf{EndIf}\\
        (E.3) Evaluate Objective and gradient\\ $y_{k+1} = f(x_{k+1})$,\\ $g_{k+1} = \nabla f(x_{k+1}).$ \\
        (E.4) Determine State and Reward \\
        $s_{k+1}=S(\{x_i, y_i,g_i\}_{i=1}^{k+1},k,\theta_S),$\\
        $r_{k+1} = R(\{x_i, y_i,g_i\}_{i=1}^{k+1},k,\theta_R).$ };
        \node [block, below of=environment] (agent)
        {\textbf{Agent} \\ 
        \textbf{If} $k\geq1$,\\
        (A.1) Update Q-Values: \\ 
        $Q(s_k, a_k) \mathrel{{+}{=}} \alpha \left( r_k + \gamma Q(s_{k+1}, a_{k+1}) - Q(s_k, a_k) \right)$

        \textbf{EndIf}\\
        \textbf{If} $(k+1)\leq K$ \\
        (A.2) Index update: $k \gets k+1$\\
        (A.3) Action selection: $a_k=\pi[Q(s_k,a)]$\\
        \textbf{EndIf}
        };

        \draw [line] (environment.east) -- ++(0.5,0) |- node [near start, left=1.0cm, below] {$s_{k+1}, r_{k+1}$} (agent.east);
        \draw [line] (agent.west) -- ++(-0.5,0) |- 
            node [near start, right=0.5cm, below] {$a_k$} (environment.west);

    \end{tikzpicture}
    \caption{Agent-Environment interaction process for Sequential Update Selection (SUS) in unconstrained first-order optimization. For $k=0,1,\ldots, K$.}
    \label{fig:AE-process2}
\end{figure}

As a next step in the transition from conventional first-order optimization to (RL)-based optimization, instead of a single fixed update function, a finite set or tuple with pairs of different possible update functions and parameters $ \mathcal{H} = \{(H_1,\theta_1) \ldots, (H_J,\theta_J)\}$ is introduced. Here each update function $H_j$ is associated with a specific parameter set $\theta_{j}$. The set $\mathcal{H}$ can represent a group of selected update functions, such as GD, Momentum, or NAG updates with various parameter settings. The update scheme in expressions \ref{eq:yk}, \ref{eq:gk},
\ref{eq:xkp1} and 
\ref{eq:xk} can be segregated and cast as the Agent-Environment interaction process model described in Fig.~\ref{fig:AE-process2}. 
In the setting of this process model, at each iteration (or process cycle) $k$,  the agent determines an action $a_k\in \mathcal{A}=\{1, \ldots, J\}$ which corresponds to an index of an element in the update set $\mathcal{H}$. 
Note that in the figure and for the remainder of this text the index notation has been modified to match the commonly used 'Sutton-Barto' \cite{Sutton98} convention, which defines all indices of environment output quantities to be incremented and indexed as "$k+1$".     

\vspace{0.3cm}
The following paragraph describes a top-level overview of the Sequential Update Selection (SUS) process model in Fig. \ref{fig:AE-process2}, while details will be discussed in the following subsection.
Similarly, as in many conventional first-order optimization algorithms, the process is initialized with an arbitrary point $x_{k+1}$ for $k=0$. In later iterations $k>0$ the action $a_k$ is applied by selecting the corresponding update function $H_{j}$ and parameters $\theta_{j}$ such that $j=a_k$. After this, the update is applied and a new design variable vector $x_{k+1}$ is determined (E.2). Then the corresponding objective function value $y_{k+1}$ and gradient $g_{k+1}$ are evaluated (E.3). The state $s_{k+1}$ and reward
$r_{k+1}$ are determined by state function $S()$ and reward function $R()$ respectively (E.4). 
When the state $s_{k+1}$ and reward $r_{k+1}$ are passed from the environment to the agent, the agent updates the estimate of the expected returns (A.1) in the Q-table $Q(s,a)$ based on the current and previous iteration. Then, the agent increments the current index update (A.2), after which a new action is selected (A.3) based on the policy $\pi[~]$, which is an operator over the Q-values. The new action $a_k$ is then forwarded to the Environment, after which the described procedure will repeat. 

\subsection{SUS Details and Implementation Examples}
\subsubsection{Action Selection}
In this setting, $\epsilon$-greedy action selection is used by the agent, such that:
\begin{equation*}
a_k=\pi^{\epsilon}(s_k) = \begin{cases} 
\text{Uniform}(\mathcal{A}) & \text{with prob. } \epsilon, \\
\arg\max_{a \in A} Q(s_k, a) & \text{with prob. } 1 - \epsilon.
\end{cases}
\end{equation*}
In a nutshell: with probability $\epsilon$ a random action is chosen, otherwise the action with the highest expected conditional return is chosen.  
\subsubsection{Q-Values, returns, and updates}
The values in the Q-table are defined as the expectation of the future returns following policy $\pi$ from state $s$ after taking action $a$.
\[
Q^\pi(s, a) = \mathbb{E}[G_t | s_k = s, a_k = a].
\]
The return $G_k$ is defined as the sum of discounted rewards, (with discount factor $\gamma$) received after iteration k. 
\[
G_k = \sum_{i=0}^{K} \gamma^{i} R_{k+i+1}
\]
These expected values of the return, in the Q-tables or Q-functions are typically not known in advance. Under mild assumptions \cite{Sutton98}, they can be initialized arbitrarily and learned during the training process using Q-learning or 
'SARSA'~\cite{Sutton98} updates 
\small
\begin{equation}
    Q(s_k, a_k) = (1 - \alpha_k) Q(s_k, a_k) + \alpha_k \left[r_{k+1} + \gamma Q(s_{k+1}, a_{k+1})\right],
\end{equation}
\normalsize
with learning rate $\alpha_k$. 
The Q-values associated with the pair $(s_k,a_k)$ can thus be updated after each training iteration when the 'future' state $s_{k+1}$ and reward $r_{k+1}$ have been determined. For partial state representations as used in the scope of this work, SARSA-updates are not guaranteed to converge to an optimal policy but can be used as a heuristic. 

\subsubsection{Concrete Action Sets}
An example of a discrete action set for update selection enhancement is a set with 2 or $J$ different update algorithms $H_j$ and their associated parameter sets, such as, the set of GD and NAG updates and their associated parameters. 
\[
\mathcal{H} = \{ (\text{GD}_1,\theta_1 ),~( \text{NAG}_2,\theta_2) \}
\]
with $\theta_1=(\eta_1)$ and $\theta_2=(\eta_2,\mu_2,\delta_2)$. Different types of useful updates are, for example, a Greedy Uniform Random Update (GURU), with lower and upper sampling bound parameters. Alternatively, action sets with the same update function but distinct associated parameter sets can be used. 
For each type of update, the associated parameters, can be arbitrary or or the result of a preliminary conventional hyperparameter optimization. 
For some algorithms such as NAG updates, it might be useful to track the previous type of executed step inside the environment to reinitialize the velocity to zero after a random step. 

\subsubsection{State Representations}
An example of a useful one-dimensional discrete state representation is a classification of the objective value at the current iteration relative to a lower bound $l$ and upper bound $u$ discretized in $m_1$ equal-sized classes. 
\[
s^1_{k+1} = \left\lfloor \min\left(1, \max\left(0, \frac{y_{k+1} - l}{u - l}\right)\right) \cdot (m_1 - 1) \right\rfloor + 1.
\]
Instead of using the objective function value directly, a logarithmic transformation of the objective function value could be useful when the objective is non-negative with a large range, as is typically the case for loss functions or residual norms.  
Another concrete example of a 1-dimensional discrete state representation is related to the fraction of iterations executed in the past, relative to the maximum function evaluation budget, discretized in $m_2$ bins.  
\[
s^2_{k+1} = \left\lfloor \frac{k}{K/m_2} \right\rfloor + 1
\]
These two one-dimensional state representations can be combined to form a simple yet useful 2-dimensional state representation
\begin{equation}
s_{k+1} 
:= (s^1_{k+1},s^2_{k+1}). \label{eq:state} 
\end{equation}
These primitive state representation concepts can be combined and extended, with state dimensions that consider properties of the gradients. In the above examples, discrete state space representations and action sets are used, such that Q-tables can be used. The concepts can, however, also be generalized to continuous and mixed representations and Q-functions. 

\subsubsection{Reward models}
One possible basic reward model for RL-based optimization is related to the relative improvement of the objective function during the last iterate. The following update scheme can be used: $r_1=0$ and
\begin{equation}
r_{k+1}=y_{k}-y_{k+1}. \label{eq:rew} 
\end{equation}
This reward strategy results in a total return equal to the overall objective improvement. For some types of non-negative objectives, such as losses or residual norms, it can be useful to emphasize the small changes in the 'end game' of the optimization run more strongly in the rewards by using a logarithmic transformation.
\[
r_{k+1}=\ln({y_{k}/y_{k+1}}).
\]
Besides or instead of these 'immediate' or step rewards, also other reward models, with delayed or terminal rewards can be used, for example, when a target or threshold is met. However, immediate rewards are typically easier to learn.  

\section{Case Studies and Results}
\subsubsection{Setup}
To investigate the described RL-based optimization approach several proof-of-concept case studies have been performed. 
The setting of these studies is iterative minimization of a class of quadratic functions
\[
f(x)=\frac{1}{2}x^TAx - b^Tx + c
\]
with $x \in \mathbb{R}^d$, symmetric positive definite matrix $A\in \mathbb{R}^{d\times d} $, with condition number $\kappa$, arbitrary $b \in \mathbb{R}^d$, and $c=1/2b^TA^{-1}b $ such that $f(x)\geq0$ for all $x$, with equality at the solution $x^*$ that coincides with the solution of the associated linear equation system $Ax^*=b$. In the setting of first-order problems only the objective function value $f(x)$ and gradient $\nabla f(x)$ are available, and the matrix $A$ is not directly observable. In the fixed budget setting the objective value is considered after a maximum number of iterations $K$. While for a fixed target $T$ setting, increments are performed until $f(x_k)\leq T$.
As a reference solution method, we use the common version of NAG with exponential decay,  
with parameters $\eta_1,\mu$ and decay $\delta$ that are hyperoptimized. 
The hyperparameter optimization is conducted over a maximum of 500 iterations using the Nelder-Mead simplex method. Each iteration involves performing optimization runs of length $K$ on 50 randomly sampled instances from a 'training set' of 100 problem instances. The performance averaged over these 50 instances serves as the objective function for the hyperparameter optimization. For the RL training and the hyperparameter optimization, the same training set drawn from the problem set $\mathcal{P}_{d,\kappa}$ is used. Where condition number $\kappa$ can be fixed or a random variable with a specified distribution. The RL training procedure involves using the epsilon-greedy policy combined with the SARSA algorithm, where both the exploration rate $\epsilon$ and the learning rate $\alpha$ decay exponentially from their initial values $\epsilon_0$ and $\alpha_0$ over increasing episodes $n = 1, 2, \ldots, N$.

To investigate and compare the performance of the presented approach the state representation and reward strategy as described in equations ~\ref{eq:state} and ~\ref{eq:rew} were used. For the studies, action and update sets $\mathcal{H}_{i\in 1,2,3}$ of only 2-4 possible updates were used:
\begin{itemize}
    \item Update set $\mathcal{H}_1=\{ \{ \text{NAG},(\eta_i,\mu^*,\delta^*)\}_{i=1}^{k=2}\}$ with: $\eta_i=[0.5\eta^*, \eta^*]$ where $\eta^*$, $\mu^*$ and $\delta^*$ are the hyperoptimized parameters of the reference baseline NAG, for the corresponding problem set $\mathcal{P}_{d,\kappa}$. 
    \item $\mathcal{H}_2=\{ \{ \text{NAG},(\eta_i,\mu^*,\delta^*)\}_{i=1}^{k=4}\}$ \newline with: $\eta_i=[0.25\eta^*, 0.5\eta^*, \eta^*, 2\eta^*]$. 
    \item $\begin{aligned}[t]
\mathcal{H}_3 = \{ & \{\text{GURU}, (-1, 1)\}, \\
                   & \{\text{GD}, (\eta_2 = 0.003)\}, \\
                   & \{\text{GD}, (\eta_3 = 0.001 \}, \\
                   & \{\text{NAG}, (\eta_4 = 0.0006, \mu_3 = 0.6, \delta = 0.0001)\} \}
\end{aligned}$

\end{itemize}



\vspace{0.1cm}
\subsubsection{Results}
\begin{figure}[t]
\centering
\includegraphics[width=\linewidth]
{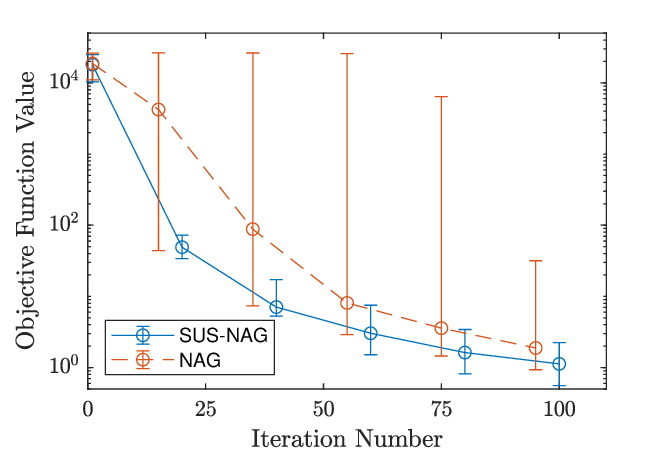}
\caption{Medians and (0.25,0.75) quantiles of objective values for successive iterations, of SUS-NAG and  Hyperoptimized-NAG, for parameters $d=100$, $K=100$, $\kappa \sim \text{Uniform}[1E2,1E3]$, $m_1=20$, $m_2=40$, $\mathcal{H}_2$. Training parameters $\epsilon_0=0.99$, $\alpha_0=0.3$, decaying to 0.5\% of their initial value at final episode 12800.}
\label{fig:ObjEvo}
\end{figure}

\begin{figure}[t]
\centering
\includegraphics[width=\linewidth]
{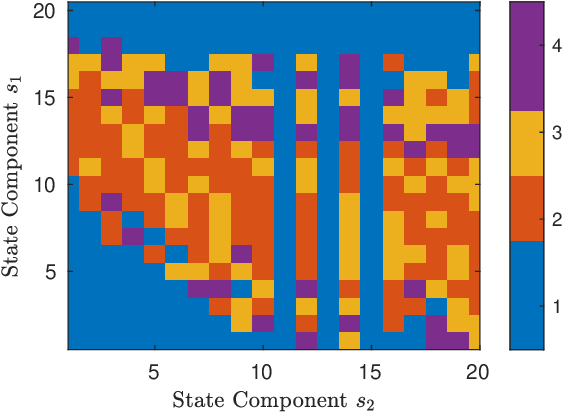}
\caption{Example of a greedy policy table, where the colors represent the optimal action index in action set $\mathcal{H}_3$ for all states. Problem parameters $d=10$, $K=20$, $\kappa \sim \text{Uniform}[1E3,1E3]$, $m_1=20$, $m_2=20$, $\mathcal{H}_3$. Training parameters $\epsilon_0=0.99$, $\alpha_0=0.3$, decaying to 0.5\% of their initial value at final episode $N=12800$.}
\label{fig:Ptable}
\end{figure}

All results are based on independent training and test sets with 100 problem instances each. A characteristic example of the best objective value history over the iteration history for the two approaches: hyperoptimized NAG, and Sequential Update Selection NAG (SUS-NAG) is shown in Fig.~\ref{fig:ObjEvo}. In the context of this minimization example with the parameters described in the caption of \ref{fig:ObjEvo}, SUS-NAG has uniformly better performance distribution statistics than conventional NAG with hyperoptimized parameters.
An illustrative example of a greedy policy table after training is given in Fig. \ref{fig:Ptable}, where for each state, the colors correspond to the update type index that, according to the SARSA training results, is expected to give the highest total returns. More closely inspecting this example of SUS Enhancement that included a random update option, resulted in a policy that contains concepts similar to commonly used warm-start and multi-start strategies. 

\begin{figure}[t]
\centering
\includegraphics[width=\linewidth]
{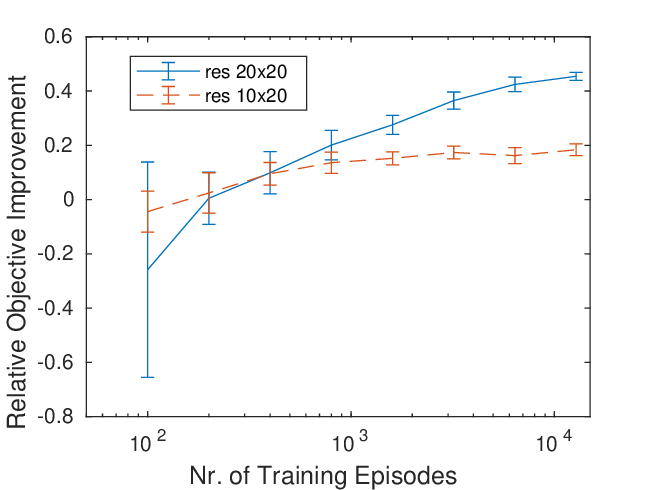}
\caption{Means and standard deviations of average relative objective value improvement: $(y_K^{\text{NAG}}-y_K^{\text{SUS}-\mathcal{H}_1})/y_K^{\text{NAG}}$, for parameters $d=100$, $K=50$, $\kappa \sim \text{Unif.}[1E2,1E3]$, $\mathcal{H}_1$, $m_1=10$, $m_2=20$, as well as $m_1=20$, $m_2=40$. Training parameters: $\epsilon_0=0.99$, $\alpha_0=0.3$, decaying to 0.5\% of their initial value at the final episode.}
\label{fig:Train}
\end{figure}

Figure \ref{fig:Train} shows means and standard deviations of the relative objective improvement in a fixed budget setting with $K=50$ optimization iterations, for an increasing number of RL-training episodes. 
The diagram in Fig. \ref{fig:Train} shows that for insufficient training length, the performance of the RL-process is less than hyperoptimized NAG, while for increasing training length performance gains increase to about 40\% of relative objective improvement. These results also confirm the intuition that higher-resolution state representations could improve performance but also require more training.   

\begin{figure}[h]
\centering
\includegraphics[width=\linewidth]{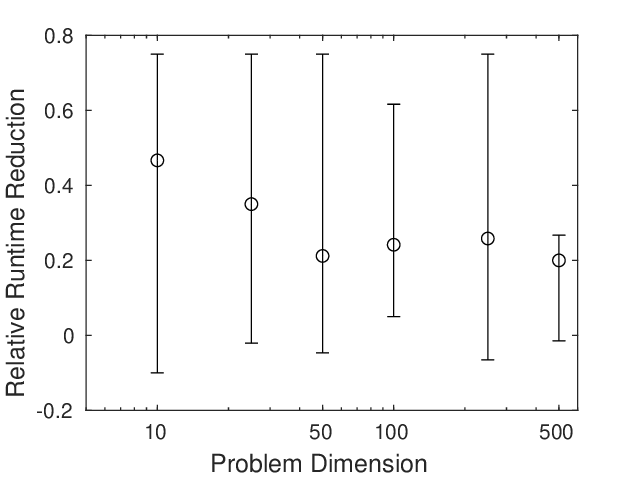}
\caption{Medians and (0.25, 0.75) quantiles of the relative run time reduction: $(k_T^{\text{NAG}}-k_T^{\text{SUS}-\mathcal{H}_2})/k_T^{\text{NAG}}$ for increasing problem dimension. Parameters: $\kappa \sim \text{Unif.}[1E2,1E3]$, $m_1=20$, $m_2=40$, $\mathcal{H}_2$. Training parameters $\epsilon_0=0.99$, $\alpha_0=0.3$, decaying to 0.5\% of their initial value at the final episode 12800.}
\label{fig:runt}
\end{figure}

From a fixed target setting parameter study, the relative runtime reduction of SUS-$\mathcal{H}_2$ to hyperoptimized NAG, for increasing problem dimension are displayed in Fig.~\ref{fig:runt}. The results indicate that for the same fixed state representation parameter settings and training parameters, the result distribution statistics mildly depend on the problem dimension with a mildly decreasing trend. All the medians of the relative runtime reduction are between 20 and 45\%. 
Overall the case study results showed potential performance benefits of SUS variants, over the hyperoptimized NAG baseline. In a less space-restricted setting, further investigations on a larger variety of problems are planned to be presented. 

\section{Conclusion, Discussion, and outlook}
In the presented proof-of-concept study an RL-based agent-environment interaction process for sequential update selection in first-order optimization was investigated. In this process model, the environment is responsible for executing function (and gradient) evaluations, while the agent dynamically selects iteration updates types based on the partial state representation and policy. In contrast to contemporary trends, this work combined elementary RL methods with simple low-dimensional state representations and relatively small action sets with primitive updates from conventional optimization algorithms in order to reduce and manage learning complexity and training effort. Despite the relative simplicity of this heuristic approach, the results from the case studies showed that better performance than theoretically optimal hyperoptimized NAG could be achieved.  

In retrospect, it seems natural that instead of using a conventional optimization algorithm such as exponential decay NAG with 3 hyperoptimized parameters, the use of an agent with a policy and action set with more parameters can outperform the former algorithm if both are trained and deployed on instances of the same problem class. Indeed, improved performance using sophisticated RL-based optimization over conventional approaches was also previously reported in the literature \cite{chen2022learning}. Complementarily, this work highlights that also using only primitive notions of states, rewards, and policies can lead to performance improvements. 
The additional algorithmic overhead and complexity for deploying the described agent-environment interactions are relatively small. However, using basic temporal difference Q-value updates combined with epsilon-greedy training requires about an order of magnitude more initial training effort compared to conventional hyperparameter optimization. When an efficient agent-based optimization strategy is widely deployed, this initial training cost might be worth the investment.


Future work can extend the presented study by exploring alternative update operators and state representations, as well as researching hierarchical agentic architectures, pre-training, and generalization across different problem classes (e.g., \cite{sala17global}).
Open challenges for further developments in RL-based optimization include finding useful and efficient state representations, reward strategies, and learning strategies.
Research in these directions aligns with the paradigm of computational rationality, which aims to identify actions with the highest expected utility while considering the costs of computation and uncertainty \cite{gershman15computational,sala22computational}. Even if the characteristics of a specified problem class are unknown but representative training instances are available, RL-based optimization could enable the alignment of effective update policies with problem class characteristics by 'training' and learning from interactions with problem instances.    

\balance






\bibliographystyle{./IEEEtran} 
\bibliography{./IEEEabrv,./IEEEexample}

\begin{thebibliography}{10}
\providecommand{\url}[1]{#1}
\csname url@rmstyle\endcsname
\providecommand{\newblock}{\relax}
\providecommand{\bibinfo}[2]{#2}
\providecommand\BIBentrySTDinterwordspacing{\spaceskip=0pt\relax}
\providecommand\BIBentryALTinterwordstretchfactor{4}
\providecommand\BIBentryALTinterwordspacing{\spaceskip=\fontdimen2\font plus
\BIBentryALTinterwordstretchfactor\fontdimen3\font minus \fontdimen4\font\relax}
\providecommand\BIBforeignlanguage[2]{{%
\expandafter\ifx\csname l@#1\endcsname\relax
\typeout{** WARNING: IEEEtran.bst: No hyphenation pattern has been}%
\typeout{** loaded for the language `#1'. Using the pattern for}%
\typeout{** the default language instead.}%
\else
\language=\csname l@#1\endcsname
\fi
#2}}

\bibitem{Nesterov1983}
Y.~E. Nesterov, ``A method of solving a convex programming problem with convergence rate o(1/k\^{}2),'' \emph{Soviet Mathematics Doklady}, vol.~27, pp. 372--376, 1983, english translation of original article in Doklady Akademii Nauk SSSR, Vol. 269, No. 3, pp. 543--547, 1983.

\bibitem{sutskever2013importance}
\BIBentryALTinterwordspacing
I.~Sutskever, J.~Martens, G.~Dahl, and G.~Hinton, ``On the importance of initialization and momentum in deep learning,'' in \emph{International conference on machine learning}.\hskip 1em plus 0.5em minus 0.4em\relax PMLR, 2013, pp. 1139--1147. [Online]. Available: \url{http://proceedings.mlr.press/v28/sutskever13-supp.pdf}
\BIBentrySTDinterwordspacing

\bibitem{wensing2023optimization}
P.~M. Wensing, M.~Posa, Y.~Hu, A.~Escande, N.~Mansard, and A.~Del~Prete, ``Optimization-based control for dynamic legged robots,'' \emph{IEEE Transactions on Robotics}, 2023.

\bibitem{oka23nesterov}
T.~Oka, R.~Misawa, and T.~Yamada, ``Nesterov’s acceleration for level set-based topology optimization using reaction-diffusion equations,'' \emph{Applied Mathematical Modelling}, vol. 120, pp. 57--78, 2023.

\bibitem{liu2017survey}
X.~Liu, P.~Lu, and B.~Pan, ``Survey of convex optimization for aerospace applications,'' \emph{Astrodynamics}, vol.~1, pp. 23--40, 2017.

\bibitem{agrawal2020learning}
A.~Agrawal, S.~Barratt, S.~Boyd, and B.~Stellato, ``Learning convex optimization control policies,'' in \emph{Learning for Dynamics and Control}.\hskip 1em plus 0.5em minus 0.4em\relax PMLR, 2020, pp. 361--373.

\bibitem{domhan2015speeding}
T.~Domhan, J.~T. Springenberg, and F.~Hutter, ``Speeding up automatic hyperparameter optimization of deep neural networks by extrapolation of learning curves,'' in \emph{Twenty-fourth international joint conference on artificial intelligence}, 2015.

\bibitem{yang2020hyperparameter}
L.~Yang and A.~Shami, ``On hyperparameter optimization of machine learning algorithms: Theory and practice,'' \emph{Neurocomputing}, vol. 415, pp. 295--316, 2020.

\bibitem{dauphin2015equilibrated}
Y.~Dauphin, H.~De~Vries, and Y.~Bengio, ``Equilibrated adaptive learning rates for non-convex optimization,'' \emph{Advances in neural information processing systems}, vol.~28, 2015.

\bibitem{guo2019new}
X.~Guo, B.~van Stein, and T.~B{\"a}ck, ``A new approach towards the combined algorithm selection and hyper-parameter optimization problem,'' in \emph{2019 IEEE Symposium Series on Computational Intelligence (SSCI)}.\hskip 1em plus 0.5em minus 0.4em\relax IEEE, 2019, pp. 2042--2049.

\bibitem{sala20benchmarking}
R.~Sala and R.~M{\"u}ller, ``Benchmarking for metaheuristic black-box optimization: perspectives and open challenges,'' in \emph{2020 IEEE Congress on Evolutionary Computation (CEC)}.\hskip 1em plus 0.5em minus 0.4em\relax IEEE, 2020.

\bibitem{meunier21black}
L.~Meunier, H.~Rakotoarison, P.~K. Wong, B.~Roziere, J.~Rapin, O.~Teytaud, A.~Moreau, and C.~Doerr, ``Black-box optimization revisited: Improving algorithm selection wizards through massive benchmarking,'' \emph{IEEE Transactions on Evolutionary Computation}, vol.~26, no.~3, pp. 490--500, 2021.

\bibitem{wu2023selecting}
Y.~Wu and L.~Liu, ``Selecting and composing learning rate policies for deep neural networks,'' \emph{ACM Transactions on Intelligent Systems and Technology}, vol.~14, no.~2, pp. 1--25, 2023.

\bibitem{polyak1964some}
B.~T. Polyak, ``Some methods of speeding up the convergence of iteration methods,'' \emph{Ussr computational mathematics and mathematical physics}, vol.~4, no.~5, pp. 1--17, 1964.

\bibitem{su2016differential}
W.~Su, S.~Boyd, and E.~J. Candes, ``A differential equation for modeling nesterov's accelerated gradient method: Theory and insights,'' \emph{Journal of Machine Learning Research}, vol.~17, no. 153, pp. 1--43, 2016.

\bibitem{sala2022euler}
R.~Sala, A.~Schlüter, C.~Sator, and R.~Müller, ``Unifying relations between iterative linear equation solvers and explicit euler approximations for associated parabolic regularized equations,'' \emph{Results in Applied Mathematics}, vol.~13, p. 100227, 2022.

\bibitem{shi2023learning}
B.~Shi, W.~Su, and M.~I. Jordan, ``On learning rates and schr{\"o}dinger operators,'' \emph{Journal of Machine Learning Research}, vol.~24, no. 379, pp. 1--53, 2023.

\bibitem{ge2019step}
R.~Ge, S.~M. Kakade, R.~Kidambi, and P.~Netrapalli, ``The step decay schedule: A near optimal, geometrically decaying learning rate procedure for least squares,'' \emph{Advances in neural information processing systems}, vol.~32, 2019.

\bibitem{smith2017cyclical}
L.~N. Smith, ``Cyclical learning rates for training neural networks,'' in \emph{2017 IEEE winter conference on applications of computer vision (WACV)}.\hskip 1em plus 0.5em minus 0.4em\relax IEEE, 2017, pp. 464--472.

\bibitem{smith2017don}
S.~L. Smith, P.-J. Kindermans, C.~Ying, and Q.~V. Le, ``Don't decay the learning rate, increase the batch size,'' \emph{arXiv preprint arXiv:1711.00489}, 2017.

\bibitem{holland1975adaptation}
J.~H. Holland, \emph{Adaptation in natural and artificial systems. an introductory analysis with applications to biology, control and artificial intelligence}.\hskip 1em plus 0.5em minus 0.4em\relax University of Michigan Press, 1975.

\bibitem{vapnik2006estimation}
V.~Vapnik, \emph{Estimation of dependences based on empirical data}.\hskip 1em plus 0.5em minus 0.4em\relax Springer Science \& Business Media, 2006.

\bibitem{bellman57dynamic}
R.~Bellman, \emph{Dynamic Programming}.\hskip 1em plus 0.5em minus 0.4em\relax Princeton, NJ: Princeton University Press, 1957.

\bibitem{Sutton98}
R.~S. Sutton and A.~G. Barto, \emph{Reinforcement Learning: An Introduction}.\hskip 1em plus 0.5em minus 0.4em\relax Cambridge, MA, USA: MIT Press, 1998.

\bibitem{vilalta2002perspective}
R.~Vilalta and Y.~Drissi, ``A perspective view and survey of meta-learning,'' \emph{Artificial intelligence review}, vol.~18, pp. 77--95, 2002.

\bibitem{vanschoren2019meta}
J.~Vanschoren, \emph{Automated machine learning: methods, systems, challenges}.\hskip 1em plus 0.5em minus 0.4em\relax Springer International Publishing, 2019, ch. 2 Meta-Learning, pp. 35--61.

\bibitem{bengio1991learning}
Y.~Bengio, S.~Bengio, and J.~Cloutier, ``Learning a synaptic learning rule,'' in \emph{IJCNN-91-Seattle International Joint Conference on Neural Networks}, vol.~2.\hskip 1em plus 0.5em minus 0.4em\relax IEEE, 1991, pp. 969--vol.

\bibitem{gregor2010learning}
K.~Gregor and Y.~LeCun, ``Learning fast approximations of sparse coding,'' in \emph{Proceedings of the 27th international conference on international conference on machine learning}, 2010, pp. 399--406.

\bibitem{zaremba2016learning}
W.~Zaremba, T.~Mikolov, A.~Joulin, and R.~Fergus, ``Learning simple algorithms from examples,'' in \emph{International conference on machine learning}.\hskip 1em plus 0.5em minus 0.4em\relax PMLR, 2016, pp. 421--429.

\bibitem{andrychowicz2016learning}
M.~Andrychowicz, M.~Denil, S.~Gomez, M.~W. Hoffman, D.~Pfau, T.~Schaul, B.~Shillingford, and N.~De~Freitas, ``Learning to learn by gradient descent by gradient descent,'' \emph{Advances in neural information processing systems}, vol.~29, 2016.

\bibitem{li2016learning}
K.~Li and J.~Malik, ``Learning to optimize,'' \emph{arXiv preprint arXiv:1606.01885}, 2016.

\bibitem{solozabal2019virtual}
R.~Solozabal, J.~Ceberio, A.~Sanchoyerto, L.~Zabala, B.~Blanco, and F.~Liberal, ``Virtual network function placement optimization with deep reinforcement learning,'' \emph{IEEE Journal on Selected Areas in Communications}, vol.~38, no.~2, pp. 292--303, 2019.

\bibitem{zhang2023multi}
B.~Zhang, W.~Hu, A.~M. Ghias, X.~Xu, and Z.~Chen, ``Multi-agent deep reinforcement learning based distributed control architecture for interconnected multi-energy microgrid energy management and optimization,'' \emph{Energy Conversion and Management}, vol. 277, 2023.

\bibitem{esmaeili2023agent}
A.~Esmaeili, Z.~Ghorrati, and E.~T. Matson, ``Agent-based collaborative random search for hyperparameter tuning and global function optimization,'' \emph{Systems}, vol.~11, no.~5, p. 228, 2023.

\bibitem{shin2022evolutionary}
D.-H. Shin, D.-H. Ko, J.-W. Han, and T.-E. Kam, ``Evolutionary reinforcement learning for automated hyperparameter optimization in eeg classification,'' in \emph{2022 10th International Winter Conference on Brain-Computer Interface (BCI)}.\hskip 1em plus 0.5em minus 0.4em\relax IEEE, 2022, pp. 1--5.

\bibitem{shen2024efficient}
Z.~Shen, H.~Yang, Y.~Li, J.~Kwok, and Q.~Yao, ``Efficient hyper-parameter optimization with cubic regularization,'' \emph{Advances in Neural Information Processing Systems}, vol.~36, 2024.

\bibitem{chen2022learning}
T.~Chen, X.~Chen, W.~Chen, H.~Heaton, J.~Liu, Z.~Wang, and W.~Yin, ``Learning to optimize: A primer and a benchmark,'' \emph{Journal of Machine Learning Research}, vol.~23, no. 189, pp. 1--59, 2022.

\bibitem{mallik2024priorband}
N.~Mallik, E.~Bergman, C.~Hvarfner, D.~Stoll, M.~Janowski, M.~Lindauer, L.~Nardi, and F.~Hutter, ``Priorband: Practical hyperparameter optimization in the age of deep learning,'' \emph{Advances in Neural Information Processing Systems}, vol.~36, 2024.

\bibitem{wang2024dp}
H.~Wang, S.~Gao, H.~Zhang, W.~Su, and M.~Shen, ``Dp-hypo: An adaptive private framework for hyperparameter optimization,'' \emph{Advances in Neural Information Processing Systems}, vol.~36, 2024.

\bibitem{wang2021reliability}
L.~Wang, B.~Ni, X.~Wang, and Z.~Li, ``Reliability-based topology optimization for heterogeneous composite structures under interval and convex mixed uncertainties,'' \emph{Applied Mathematical Modelling}, vol.~99, pp. 628--652, 2021.

\bibitem{li24settling}
G.~Li, L.~Shi, Y.~Chen, Y.~Chi, and Y.~Wei, ``Settling the sample complexity of model-based offline reinforcement learning,'' \emph{The Annals of Statistics}, vol.~52, no.~1, pp. 233--260, 2024.

\bibitem{sala17global}
R.~Sala, N.~Baldanzini, and M.~Pierini, ``Global optimization test problems based on random field composition,'' \emph{Optimization Letters}, vol.~11, pp. 699--713, 2017.

\bibitem{gershman15computational}
S.~J. Gershman, E.~J. Horvitz, and J.~B. Tenenbaum, ``Computational rationality: A converging paradigm for intelligence in brains, minds, and machines,'' \emph{Science}, vol. 349, no. 6245, pp. 273--278, 2015.

\bibitem{sala22computational}
R.~Sala, ``Computational rational engineering and development: Synergies and opportunities,'' in \emph{Intelligent Systems and Applications: Proceedings of the 2021 Intelligent Systems Conference (IntelliSys) Volume 1}.\hskip 1em plus 0.5em minus 0.4em\relax Springer, 2022, pp. 744--763.

\end{thebibliography}

\end{document}